\PassOptionsToPackage{table}{xcolor}
\documentclass[runningheads]{llncs}

\usepackage{eccv}

\usepackage{eccvabbrv}

\usepackage{graphicx}
\usepackage{booktabs}

\usepackage[accsupp]{axessibility}  
\usepackage{multirow}
\usepackage{amsmath}
\usepackage{caption}
\usepackage{colortbl}
\usepackage{bm}
\usepackage{subcaption}
\usepackage{pifont}

\newcommand{\boldcheckmark}{\textbf{\ding{52}}}
\definecolor{aliceblue}{RGB}{240,248,255}
\definecolor{lightgray}{rgb}{0.83, 0.83, 0.83}

\usepackage[pagebackref,breaklinks,colorlinks]{hyperref}

\usepackage{orcidlink}

\begin{document}

\title{Uncertainty-aware Sign Language Video Retrieval with Probability Distribution Modeling} 

\titlerunning{UPRet}


\author{Xuan Wu*\inst{1} \and
Hongxiang Li*\inst{2} \and
Yuanjiang Luo\inst{3}\and
Xuxin Cheng\inst{2}\and
Xianwei Zhuang\inst{2}\and
Meng Cao\inst{4}\and
Keren Fu$^\dag$\inst{1}
}

\authorrunning{X.Wu et al.}


\institute{College of Computer Science, Sichuan University\and Peking University\and National Key Laboratory of Fundamental Science on Synthetic Vision,
Sichuan University\and Mohamed bin Zayed University of Artificial Intelligence\\
\email{2023223040230@stu.scu.edu.cn,\\ fkrsuper@scu.edu.cn}}

\maketitle
\let\thefootnote\relax\footnotetext{$^*$ These authors contributed equally to this work.}
\let\thefootnote\relax\footnotetext{ $^\dag$ Corresponding Author}

\begin{abstract}
Sign language video retrieval plays a key role in facilitating information access for the deaf community. Despite significant advances in video-text retrieval, the complexity and inherent uncertainty of sign language preclude the direct application of these techniques. Previous methods achieve the mapping between sign language video and text through fine-grained modal alignment. However, due to the scarcity of fine-grained annotation, the uncertainty inherent in sign language video is underestimated, limiting the further development of sign language retrieval tasks. To address this challenge, we propose a novel \textbf{U}ncertainty-aware \textbf{P}robability Distribution \textbf{Ret}rieval (\textbf{UPRet}), that conceptualizes the mapping process of sign language video and text in terms of probability distributions, explores their potential interrelationships, and enables flexible mappings. Experiments on three benchmarks demonstrate the effectiveness of our method, which achieves state-of-the-art results on How2Sign (59.1\%), PHOENIX-2014T (72.0\%), and CSL-Daily (78.4\%).
\end{abstract}

\section{Introduction}
\label{sec:intro}
Sign language is crucial in non-verbal communication as the primary mode of communication for deaf and hard-of-hearing people. A deep understanding of sign language videos ensures that community members have equal access to vital information, quality education, and essential services. To date, the researchers~\cite{wei2023improving,zhang2023c2st,lee2023human,jiao2023cosign,zhou2023gloss,yao2023sign} have made significant progress in advancing sign language understanding, particularly in the sign language recognition and translation. However, sign language video retrieval is an area that has yet to be fully explored. In addition, the scarcity of sign language data further limits the development of related tasks. Therefore, the development of efficient sign language video retrieval technology is urgently needed. Such advances will provide the deaf community with a more autonomous way of accessing information. Still, they will also help to alleviate the scarcity of sign language resources, thus contributing to the rich development and technological advancement of related fields.

Compared to general video, sign language video has a more nuanced visual representation where subtle changes in gestures play a key role in conveying the message. As a highly continuous visual signal, sign language is so context-dependent that it is infeasible to understand sign language video clips in isolation from the whole video. Sign language video not only presents visual information, it is also a language with its own unique grammatical rules. This leads to a more complex alignment between video and text. Therefore, sign language video-text alignment suffers from uncertainty.

\begin{figure}[t]
\centering
\includegraphics[width=0.9\linewidth]{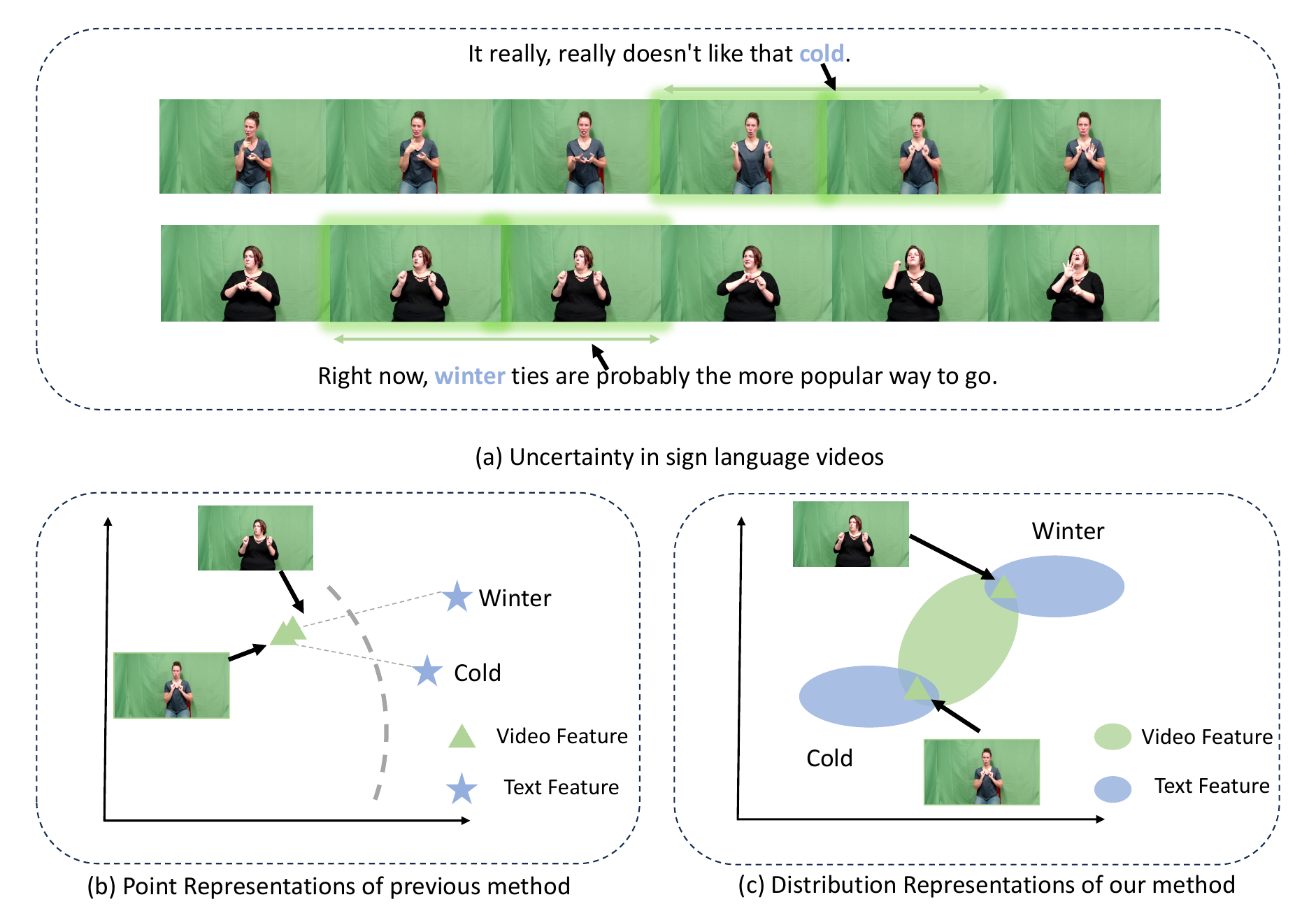}
\vspace{-1em}
\caption{
(a) Illustration of the uncertainty between sign language video and text. (b) Previous methods obtain single-point representations through one-to-one mappings, which make it difficult to capture one-to-many relationships in semantic space and thus present challenges in dealing with uncertainty in sign language scenarios. (c) Our method re-models representations in terms of probability distributions to better deal with uncertainty.
}
\label{fig:intro}
\vspace{-2em}
\end{figure}

As shown in Figure~\ref{fig:intro}(a), we show the uncertainty inherent in text matching for sign language video. (1) From the visual perspective, the polysemic nature of sign language means that the same gesture can have diverse meanings in different contexts. 
For example, the words \textit{"cold"} and \textit{"winter"} share the same visual appearance in Figure~\ref{fig:intro}(a).
This illustrates how a single gesture in sign language can have multiple interpretations, which vary depending on the context in which it is used. (2) From the textual perspective, one word may correspond to multiple sign language expressions due to the inherent complexity of natural language, such as synonymy and polysemy. Moreover, dialectal and geographical variations also bring challenges to sign language comprehension.

Existing works~\cite{duarte2022sign,cheng2023cico} have limitations in accurately matching subtle semantic differences between sign language videos and user queries, \eg, CiCo~\cite{cheng2023cico} identifies fine-grained mappings between sign language and natural language through cross-lingual contrastive learning. Such a one-to-one mapping fails to capture the polysemy and complexity of sign languages, as shown in Figure~\ref{fig:intro}(b). Due to the high cost of manual annotation, the labels obtained often cannot cover the exact correspondence of each frame to each word. Therefore, these limitations significantly hinder previous method~\cite{cheng2023cico} from achieving fine-grained alignments from coarse annotated data, limiting the depth of understanding between sign language videos and textual queries.
To capture as comprehensive a representation of sign language as possible from a limited number of annotations, We have to explore other learning signals to improve vanilla contrastive learning.
Therefore, we innovatively introduce a distributional modeling perspective to comprehensively capture the correspondence between sign language videos and texts through dynamic semantic alignment. As shown in Figure~\ref{fig:intro}(c), in contrast to previous work~\cite{cheng2023cico}, which treats sign language and natural language as point representations, we model sign language video and corresponding text as multivariate Gaussian distribution. This method allows us to explore the correspondence between video and text within a wider semantic space, thus capturing the uncertainty and polysemy of sign language more accurately.

To this end, we propose a new \textbf{U}ncertainty-aware \textbf{P}robability Distribution \textbf{Ret}rieval (\textbf{UPRet}), leverage probability distributions reconstruct the representations between sign language video and text, explore their respective possibility spaces, and consider sign language video retrieval as a process of probability distribution matching. Specifically, we effectively capture the uncertainty inherent in sign language video and text by modeling each as a probability distribution. This allows for a more flexible manner of one-to-many matching between sign language video and text. Moreover, through Monte Carlo sampling, we explored in depth the structure and associations of the data distribution. We achieve fine-grained cross-modal alignment by optimal transport at the minimum cost between computationally distributed transformations, providing a novel perspective on sign language retrieval.
Experimental results on three sign language video-text retrieval benchmark datasets (\eg, How2Sign~\cite{duarte2021how2sign}, PHOENIX-2014T~\cite{camgoz2018neural}, CSL-daily~\cite{zhou2021improving}) show the advantages of the proposed methods.

Our contributions are summarized as follows:
\begin{itemize}
    \item We provide a new perspective to consider video and text as probability distributions to capture the uncertainty.
    \item We introduce Optimal Transport (OT) as a fine-grained measure of the distance between distributions.
    \item Experiments demonstrate that our method achieves state-of-the-art performance on three sign video-text benchmarks. 
\end{itemize}

\section{Related Work}

\subsection{Sign Language Retrieval}
With a better understanding of the needs of the deaf community, sign language video understanding has become a key tool to support accessible information access and communication. Technological advances in sign language recognition and translation are designed to help deaf users access and understand video content. The goal of sign language video recognition is divided into two steps. Firstly, it aims to accurately recognize the gestures displayed in the video. Secondly, these gestures need to be categorized into pre-defined sign language words or phrases. This field is divided into two subcategories: isolated sign language recognition (ISLR) and continuous sign language recognition (CSLR). ISLR~\cite{zuo2023natural,li2020word,jiang2021skeleton,li2020transferring} focuses on recognizing individual gestures in video and correctly matching them to their corresponding sign language symbols. CSLR~\cite{cheng2020fully,koller2017re,zuo2022c2slr,camgoz2020sign,zheng2023cvt,wei2023improving} aims to convert a sequence of sign language movements into a sequence of text, explaining the meaning word by word in a sign language video. Unlike recognition tasks~\cite{cihan2017subunets,huang2018video,cui2017recurrent}, sign language translation~\cite{chen2022two,li2020tspnet,camgoz2020multi,camgoz2020sign,zhou2023gloss,yao2023sign} takes into account the unique syntax of sign language, which requires word order to be adjusted during transcription to ensure translation accuracy. Sign language translation goes far beyond the simple translation of visual signals into words, digging deeper into the hidden meanings of sign language, such as facial expressions and body movements. The goal is to fully capture the essence of sign language and translate it into equivalent expressions that match the grammar and cultural traditions of the target language.

Building on the foundation laid by sign language recognition and translation, which significantly improves content accessibility, sign language video retrieval emerges as a critical next step. It involves more than just recognizing and understanding sign language; it also requires the ability to quickly and accurately identify sign language segments within large datasets that match textual queries. SPOT-ALGN~\cite{duarte2022sign}, a groundbreaking study, directly defines sign language retrieval as a video-text retrieval task by learning cross-modal embedding and alignment of sign language video and text. CiCo~\cite{cheng2023cico} made great progress by complementing the linguistic properties of the sign language video themselves, learning sign-to-word mappings, and joining the two tasks of text-video retrieval and cross-language retrieval. However, this single-point mapping method makes it difficult to resolve the uncertainty in the two modalities of sign language video and text. We introduce more appropriate modeling of one-to-many, thus resolving the uncertainty and surpassing previous methods.

\subsection{Text-Video Retrieval}
With the remarkable achievements of the large-scale image-text pre-training model CLIP~\cite{radford2021learning}, many works~\cite{luo2022clip4clip,fang2021clip2video,zhao2022centerclip,ma2022x,xue2022clip,chen2023tagging} have begun to explore how pre-trained CLIP can be applied to video-text pairings to achieve cross-modal mapping to a unified representation space. CLIP4Clip~\cite{luo2022clip4clip} and CLIP2Video~\cite{fang2021clip2video} successfully transfer deep knowledge from CLIP to video text retrieval tasks through an end-to-end manner. Considering semantic consistency, subsequent works have been dedicated to cross-modal interactions in a specific granularity,\eg, coarse-grained level~\cite{luo2022clip4clip,gorti2022x}, fine-grained level~\cite{wang2022disentangled,fang2023uatvr} and hierarchical level interactions~\cite{wu2023cap4video,wang2023unified,jin2023video}. However, coarse-grained interactions ignore detailed gestures and actions in sign language video, while hierarchical-level interactions introduce more complex semantic information. As sign language is a highly complex and detailed visual language, it requires a precise understanding of the actual meaning of each frame and each word. Therefore, our work aims to explore the implementation of sign language video retrieval under fine-grained mapping.

Inspired by FILIP~\cite{yao2021filip}, which uses token-wise similarity between visual patches and textual words to guide the training process. Previous works~\cite{ma2022x,wang2022disentangled} implement token-wise interactions and perform fine-grained matching in the video text retrieval. To move a step further, we innovatively introduce OT. By calculating the minimum transport cost of the distribution transformation, we achieve a comprehensive assessment of the similarity of different dimensions in the feature space. This method not only matches sign language video and text that are semantically similar overall but also perceives specific actions or minor transformations in the sign language and aligns them with specific words in the text. Consequently, our method is uniquely suited to deal with the high levels of uncertainty and polysemy present in both sign language video and text.

\subsection{Probabilistic Representations}
Probabilistic representations, when first introduced using Gaussian embeddings to represent words~\cite{vilnis2014word}, skilfully captured the complex linguistic structures inherent in words. Subsequently, HBI~\cite{oh2018modeling} further extends this to images, aiming to address the inherent uncertainty in image representations and to be able to efficiently handle one-to-many problems in metric learning. They have not only been successfully applied to tasks such as face recognition~\cite{chang2020data,shi2019probabilistic}, and pose estimation~\cite{sun2020view} but have also solved problems in cross-modal scenarios. PCME~\cite{chun2021probabilistic} addresses the one-to-many challenge in image-text retrieval by learning probabilistic cross-modal joint embeddings between images and captions. Inspired by this, UATVR~\cite{fang2023uatvr} extends the application from images to videos by introducing multi-instance contrast to better handle the complex semantics and uncertainty of video, providing a more suitable method for one-to-many modeling. Although previous methods are effective in resolving uncertainty, they still have limitations in capturing deeper semantic relationships and accurately aligning data across modalities. Therefore, we stay in the context of probability distributions and introduce OT to measure and minimize the differences between the different distributions. By minimizing transport costs, we can find the best matching strategy across modalities at a global level, leading to more accurate fine-grained alignment.

\section{Method}
\subsection{Preliminary}
For a video \textit{V} and its corresponding textual description \textit{T}, sign language video retrieval aims to create a model capable of understanding and mapping between sign language video and text using cross-modal representation learning. This task consists of two main aspects: text-to-sign language video retrieval (T2V) and sign language video-to-text retrieval (V2T). For T2V, the goal is to find the most matching sign language video $v \in V$ based on the textual query $t^q$. In contrast, V2T requires the model to retrieve the most relevant textual description $t \in T$ using the sign language video as the query $v^q$.
Formally, given a pair of video \textit{V} and text\textit{T} , the features are fed into the visual encoder $F_v$ and the text encoder $F_t$ respectively to get their corresponding feature embeddings$V_i' = [v_i^1,v_i^2,\ldots v_i^{N_v}]$ and $T_i = [t_i^0,v_i^1,\ldots v_i^{N_t}]$ in Eq.\ref{eq:encoder}, Where $N_v$ and $N_t$ are denoted as the number of frames in the video and the number of words in the text. Typically, the whole video is represented by the average pooling of features across all frames, and the sentence is represented by the first [CSL] word.

\begin{gather}
     V_i' = F_v(V),\;T_i' = F_t(T), 
     \label{eq:encoder} \\
     F_i = \frac{1}{N_v}\sum_{i=1}^{N_v}v_i,\;F_t = t_i^0,\; \mathcal{S}_{\boldsymbol{v},\boldsymbol{t}}=\frac{F_t^\top F_v}{\left\|F_t\right\|\left\|F_v\right\|}.
\end{gather}
The similarity $\mathcal{S}(t,v)$ of the text-video is calculated as the inner production of $F_v$,$F_t$. Then the cross-modal contrastive loss can be formulated as:
\begin{gather}
    \mathcal{L}_{t2v}=-\frac{1}{B}\sum_{i}^{B}\log\frac{\exp(s(\mathbf{t}_i,\mathbf{v}_i))}{\sum_{j=1}^{B}\exp(s(\mathbf{t}_i,\mathbf{v}_j))}, \\
    \mathcal{L}_{v2t}=-\frac{1}{B}\sum_{i}^{B}\log\frac{\exp(s(\mathbf{v}_i,\mathbf{t}_i))}{\sum_{j=1}^{B}\exp(s(\mathbf{v}_i,\mathbf{t}_j))},\\
    \mathcal{L} = \mathcal{L}_{t2v} + \mathcal{L}_{v2t}.
\end{gather}
\subsection{Encoder}
Recent advances~\cite{momeni2020watch,varol2021read} in the field of sign-spotting have greatly facilitated the creation of large-scale sign language datasets. Previous research~\cite{duarte2022sign} has shown that sign language encoders obtained by training with large-scale sign-spotting data can be effectively used for downstream tasks. Following CiCo~\cite{cheng2023cico} uses the I3D network trained on BSL-1k~\cite{varol2021read} as domain-agnostic sign encoder $h_\xi$. To solve the problem of domain differences between BSL-1K and the target task dataset, a domain-aware sign encoder $h_\theta$ is obtained by fine-tuning the $h_\xi$ of the same architecture by pseudo-labeling on the target dataset.
The final sign encoder is a weighted combination of the two encoders.
\begin{align}
    H(v)=\alpha h_\xi(v)+(1-\alpha)h_\theta(v).
\end{align}
To obtain visual and textual features separately, we take the sign language video v and first send it to the sign encoder to extract visual features. Leveraging the powerful representational capabilities of CLIP, the output of the sign encoder is then sent to Clip's image encoder $F$, to get the final sign language video features $V = F(H(v)) \in \mathbb{R}^{N_v \times D} $, where $N_v$ represents the number of frames of the sign language, and D represents the number of dimensions. The text t is converted to lower case and fed into the text encoder $G$ to get word features $T = G(T) \in \mathbb{R}^{N_t \times D}$, where $N_t$ represents word number.

\begin{figure}[t]
\centering
\includegraphics[width=1.0\linewidth]{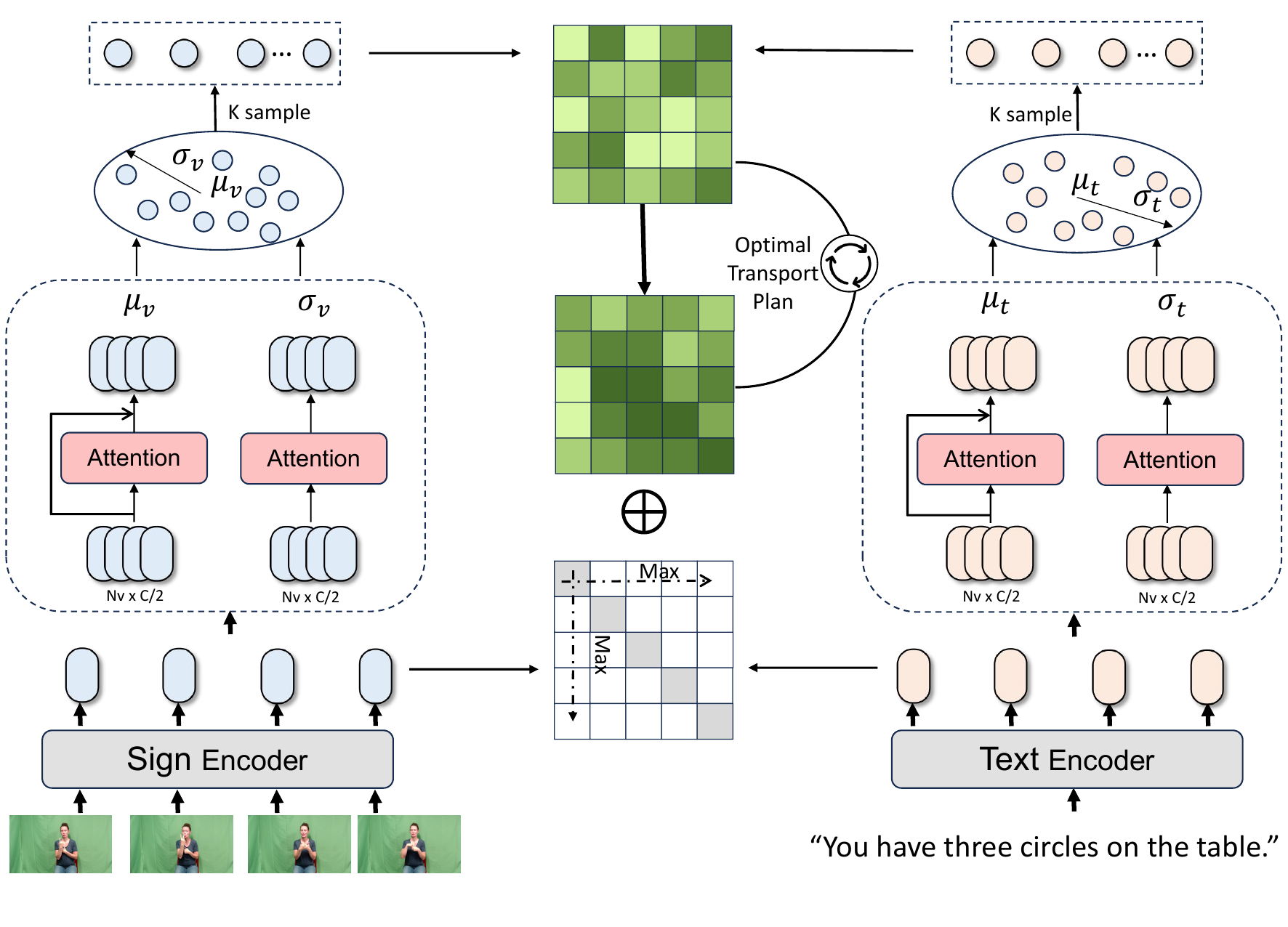}
\vspace{-2em}
\caption{
\textbf{Overview of UPRet}. Sign language video features and text features are extracted by the sign encoder and text encoder, and subsequently, we model the video distribution and text distribution. Finally, after randomly sampling the distribution, the distance of the distribution is measured using optimal transport to encourage fine-grained alignment.
}
\label{fig:pipeline}
\vspace{-1em}
\end{figure}

\subsection{Distribution Modeling}
Current feature extraction methods produce representations that correspond to individual sample points, limiting them to 1-to-1 mappings. As discussed in the Sec~\ref{sec:intro}, the polysemous nature of sign language where the same gesture may convey different meanings across various contexts. A single-point representation cannot capture the richness of the sign language. So we introduce UPRet, a sophisticated representation method that facilitates one-to-many relational mapping for real-world sign language retrieval scenarios, as shown in Figure~\ref{fig:pipeline}.

In particular, the Gaussian distribution is often used as a modeling tool to represent uncertainty in representation space~\cite{vilnis2014word,yu2019robust}. By modeling the data, the Gaussian distribution can capture the inherent complexity and variability of the data, allowing it to be effectively adapted to the characteristics of specific types of data, such as sign language. The mean vector represents the center position of distributions, and the variance vector expresses the scope of distributions. Given input features $F \in \mathbb{R}^D$,we split it into $F_{\mu}\in \mathbb{R}^{D/2}$ and $F_{\sigma}\in \mathbb{R}^{D/2}$,and utilize the multi-head attention to predict the mean vector $\mu$ and variance vector $\sigma^2$ :
\begin{align}
    &[Q_{\mu},K_{\mu},V_{\mu}] = F_{\mu}W_{qkv}\\
    &\mu = F_{\mu} + \text{ML}P(\text{softmax}(\frac{Q_{\mu}K_{\mu}^T}{\sqrt{d_k}}V_{\mu})),\\
    &[Q_{\sigma},K_{\sigma},V_{\sigma}] = F_{\sigma}W_{qkv}\\
    &\sigma = \text{MLP}(\text{softmax}(\frac{Q_{\sigma}K_{\sigma}^T}{\sqrt{d_k}}V_{\sigma})).
\end{align}
Sign language video and text will be represented by the mean to represent their maximum probability of meaning, while the variance captures the potential range of variation, \ie, uncertainty, around this meaning. After modeling the video and text features of the sign language with multivariate Gaussian distributions, we used Monte Carlo sampling~\cite{chun2021probabilistic,oh2018modeling} to explore the latent structures and relationships within these distributions. By estimating the properties of the distribution through repeated random sampling, we expect to model the multiple meanings expressed by the same gesture in different contexts. We determine the sample centrality using the mean of the distribution, while the variance highlights the diversity of the features. This approach effectively captures the uncertainty in gesture expressions. For feature refinement, K distributions are sampled from the model, with reparameterization~\cite{kingma2013auto} and average pooling used to closely simulate input features. It is defined as follows:

\begin{align}
    &f_{v_i}=\frac1{K+1}(v_i+\sum_{k=1}^K(\mu_v^i+\epsilon_k\sigma_v^i)),\\
    &f_{t_i}=\frac1{K+1}(t_i+\sum_{k=1}^K(\mu_t^i+\epsilon_k\sigma_t^i)).
\end{align}

\subsection{Optimal Transport}
In the Optimal Transport (OT) problem, we consider how to transform one probability distribution into another while minimizing the cost function. Specifically, suppose a set of source domains or suppliers $U = {u_i,i=1,2,\ldots m}$ that need to deliver goods to a set of target domains or demanders $V = {v_j|j=1,2,\ldots n}$, where $u_i$ denotes the unit of supply of supplier $i$ and $v_j$ denotes the demand of demander $j$. The unit cost of transport from supplier $i$ to demander $j$ is defined as $c_{ij}$. The goal of OT is to find a transport plan $T={T_{ij}|i=1,2,...m,j=1,2,\ldots n}$ such that the total cost of transporting the goods from the supplier to the consumer is minimized. Thus, the OT problem can be formulated as follows:
\begin{align}
    \text{minimize}\quad&\sum_{i=1}^m\sum_{j=1}^nT_{ij}c_{ij},
\end{align}
\vspace{-15pt}
\begin{align}
    \text{subject to}\quad&\sum_{i=1}^nT_{ij}=u_i,\quad
    \sum_{i=1}^mT_{ij}=v_j,\quad
    \sum_{i=1}^mu_i=\sum_{j=1}^nv_j.
\end{align}
We need to satisfy the constraints between suppliers and demanders, specifically, for the transport plan $T_{ij}$, which represents the flow of goods from supplier $i$ to demander $j$, it must be non-negative. To satisfy the supply-demand balance, the total quantity of goods supplied by each supplier must be equal to the quantity demanded by all demanders. In addition, the total quantity of goods from all suppliers needs to be equal to the total quantity demanded by all demanders. Since computing the cost matrix and solving the corresponding linear programming problem are computationally expensive, we introduce the Sinkhorn~\cite{cuturi2013sinkhorn} iterative method to improve the computational process of OT by entropy regularisation, allowing it to more accurately match and align cross-modal data.

\subsection{Sign Language Retrieval As Optimal Transport Problem}
We treat sign video and text as two probability distributions and use OT to measure the distance between these two distributions, unlike previous methods~\cite{oh2018modeling,fang2023uatvr} that rely on contrastive loss for distribution alignment. With OT, we can quantify the minimum cost required to transform one distribution into another, providing a powerful mathematical tool for the measurement of similarity between distributions. OT not only accurately captures the subtle differences and similarities between sign language video and text description, but also effectively addresses the diversity inherent in sign language expression. By comprehensively considering the distribution of sign language features and their contexts, we introduce OT to facilitate fine-grained cross-modal alignment, where the probability distribution of each frame and word is the corresponding probability score.
OT aims to find a transportation plan $\boldsymbol{T}$ minimizing the transport cost between the video distribution $\mu_v$ and the text distribution $\mu_t$. $\mu_v$ and $\mu_t$ can be formulated as:
\begin{equation}
    \mu_{v}=\sum_{i=1}^{N_{v}} p_{i}^{v} \delta\left(v_{i}\right), \quad \mu_{t}=\sum_{i=1}^{N_{t}} p_{i}^{t} \delta\left(t_{i}\right),
\end{equation}
where $\delta(\cdot)$ is a Dirac delta function placed at the input feature in the embedding space, $p_i^v$ and $p_i^t$ are the discrete probability vectors that sum to 1. Each feature has an equal probability value, \ie, $p_i^v\!=\!1/{N_v},p_i^t\!=\!1/{N_t}$. 
We utilize the token-wise alignment matrix $\mathcal{C}$ as the cost matrix $\boldsymbol{C}$, \ie, $\boldsymbol{C}\!=\!1-\mathcal{C}$. Then, The optimization problem of OT is formulated as:
\begin{gather}
\boldsymbol{T}^*=\underset{\boldsymbol{T}}{\operatorname{minimize}}\sum_{i=1}^{N_v} \sum_{j=1}^{N_t} \boldsymbol{T}_{i, j} \boldsymbol{C}_{i, j}, \\
\mathrm{s.t.~}\boldsymbol{T}\mathbf{1}_{N_t}=\mu_t,\boldsymbol{T}^\top\mathbf{1}_{N_v}=\mu_v,
\end{gather}
where $\boldsymbol{T}^*$ is the OT plan. As directly optimizing the above objective is always time-consuming, we resort to the Sinkhorn-Knopp algorithm for fast optimization using an entropy constraint. The optimization problem with a Lagrange multiplier of the entropy constraint is:
\begin{gather}
\boldsymbol{T}^*=\underset{\boldsymbol{T}}{\operatorname{minimize}}\sum_{i=1}^{N_v} \sum_{j=1}^{N_t} \boldsymbol{T}_{i, j} \boldsymbol{E}_{i, j}+\eta H(\boldsymbol{T}), \\
\mathrm{s.t.~}\boldsymbol{T}\mathbf{1}_{N_t}=\mu_t,\boldsymbol{T}^\top\mathbf{1}_{N_v}=\mu_v,
\end{gather}
where $H(\boldsymbol{T})=\sum_{ij}\boldsymbol{T}_{ij}(\log \boldsymbol{T}_{ij}-1)$ is the negative entropy regularization and $\eta\ge0$ is a regularization hyper-parameter. Then we can have a fast optimization solution with a few iterations as:
\begin{equation}
\boldsymbol{T}^*=\operatorname{diag}(\mu_t^{i})\exp(-\boldsymbol{E}/\eta)\mathrm{diag}(\mu_v^{i}),
\end{equation}
where $i$ denotes the iteration. We represent the similarity between video and text as the OT cost between the two sets of vectors. The OT distance $\mathcal{D}_{\boldsymbol{v},\boldsymbol{t}}$ can be formulated as:  
\begin{align}
\mathcal{D}_{\boldsymbol{v},\boldsymbol{t}}=\sum_{i=1}^{N_v} \sum_{j=1}^{N_t} \boldsymbol{T}_{i, j}^* (1-\boldsymbol{E}_{i, j}),
\end{align}
Given the OT distance between $\boldsymbol{v}$ and $\boldsymbol{t}$, we fix the OT plan $\boldsymbol{T}^*$ to optimize the model with cross-entropy:
\begin{gather}
    \mathcal{L}_{\mathcal{D}}=-\frac{1}{B}\sum_{i=1}^{B} \sum_{j=1}^{B}\mathrm{log}\frac{\exp\left(\mathcal{D}_{\boldsymbol{v_i},\boldsymbol{t_j}}\right)/\tau}{\sum_{i\ne j }^B\exp\left(\mathcal{D}_{\boldsymbol{v_i},\boldsymbol{t_j}} \right)/\tau}.
\end{gather}

\subsection{Training Objective}
Given sign features $V_i \in \mathbb{R}^{N_v \times D}$ of $v_i$,and word features $T_j \in \mathbb{R}^{N_t \times D}$ of $t_j$, the alignment matrix is defined as:$A=[a_{ij}]^{N_v \times N_t}$,where $a_{ij}=V_i \cdot T_j$. The cross-modal similarity can be defined as:
\begin{gather}
    \omega_{v}^{N_v} = \text{softmax}(\text{MLP}(V)),\quad \omega_{t}^{N_t} = \text{softmax}(\text{MLP}(T)), \\
    \mathcal{S}_{v,t} = \frac{1}{2}\left(\left(\sum_{i=1}^{N_v} \omega_{v}^{i} \max_{j} a_{ij} + \lambda_{ot} S_{ot}\right) + \left(\sum_{j=1}^{N_t} \omega_{t}^{j} \max_{i} a_{ij} + \lambda_{ot} S_{ot}\right)\right),
\end{gather}

where$\lambda_{ot}$ is trade-off hyper-parameters,  $\mathcal{S}_{ot}$ is calculated only during training. The full video-text contrastive loss can be formulated as:
\begin{align}
\mathcal{L}_{\mathcal{S}} =-\frac12\big[\frac1B\sum_{i=1}^B\log\frac{\exp\left(\mathcal{S}_{\boldsymbol{v}_i,\boldsymbol{t}_i}\right)/\tau}{\sum_{j=1}^B\exp\left(\mathcal{S}_{\boldsymbol{v}_i,\boldsymbol{t}_j}\right)/\tau} \\
+\frac{1}{B}\sum_{i=1}^{B}\log\frac{\exp\left(\mathcal{S}_{\boldsymbol{v}_{i},\boldsymbol{t}_{i}}\right)/\tau}{\sum_{j=1}^{B}\exp\left(\mathcal{S}_{\boldsymbol{v}_{j},\boldsymbol{t}_{i}}\right)/\tau}],     
\end{align}

\begin{table}[t]
\scriptsize
\centering
\caption{\textbf{Comparison with the SOTA on How2Sign dataset.} ``$\uparrow$'' denotes that higher is better. ``$\downarrow$'' denotes that lower is better. * denotes our re-implementation of baselines.}
\vspace{-10pt}
\begin{tabular}{lcccccccccc}
\toprule
\multirow{2}{*}{Model} & \multicolumn{5}{c}{T2V} & \multicolumn{5}{c}{V2T} \\ \cmidrule(lr){2-6} \cmidrule(lr){7-11}
 & R@1$\uparrow$ & R@5$\uparrow$ & R@10$\uparrow$ & MedR$\downarrow$ & MnR$\downarrow$ & R@1$\uparrow$ & R@5$\uparrow$ & R@10$\uparrow$ & MedR$\downarrow$ & MnR$\downarrow$ \\  \midrule
SA-SR~\cite{duarte2022sign}   & 18.9 & 32.1 & 36.5 & 62.0 & - & 11.6 & 27.4 & 32.5 & 69.0 &-  \\
SA-CM~\cite{duarte2022sign}   & 24.3 & 40.7 & 46.5 & 16.0 & - & 17.9 & 40.1 & 46.9 & 14.0 & - \\
SA-COMB~\cite{duarte2022sign} & 34.2 & 48.0 & 52.6 & 8.0  & - & 23.6 & 47.0 & 53.0 & 7.5  &-  \\
CiCo~\cite{cheng2023cico} & 56.6 & 69.9 & 74.7 & \textbf{1.0} & - & 51.6 & 64.8 & 70.1 &\textbf{1.0} & - \\
CiCo*~\cite{cheng2023cico} & 56.4 & 69.4 & 74.1 & \textbf{1.0} & 58.8 & 50.3 & 63.6 & 69.3 &\textbf{1.0} & 79.5 \\
\midrule
\rowcolor{aliceblue}
\textbf{UPRet} & \textbf{59.1} & \textbf{71.5} & \textbf{75.7} & \textbf{1.0} & \textbf{54.4} & \textbf{53.4} & \textbf{65.4} & \textbf{70} & \textbf{1.0} & \textbf{76.4} \\
\bottomrule
\end{tabular}
\label{tab:comparison sota h2s}
\end{table}

\begin{table}[t]
\scriptsize
\centering
\caption{\textbf{Comparison with the SOTA on PHOENIX-2014T dataset.} ``$\uparrow$'' denotes that higher is better. ``$\downarrow$'' denotes that lower is better. * denotes our re-implementation of baselines.}
\vspace{-10pt}
\begin{tabular}{lcccccccccc}
\toprule
\multirow{2}{*}{Model} & \multicolumn{5}{c}{T2V} & \multicolumn{5}{c}{V2T} \\ \cmidrule(lr){2-6} \cmidrule(lr){7-11}
 & R@1$\uparrow$ & R@5$\uparrow$ & R@10$\uparrow$ & MedR$\downarrow$ & MnR$\downarrow$ & R@1$\uparrow$ & R@5$\uparrow$ & R@10$\uparrow$ & MedR$\downarrow$ & MnR$\downarrow$ \\  \midrule
Translation~\cite{camgoz2020sign}& 30.2 & 53.1 & 63.4 & 4.5 & - & 28.8 & 52.0 & 60.8 &56.1 & - \\
SA-CM~\cite{duarte2022sign} & 48.6 & 76.5 & 84.6 & 2.0 & - & 50.3 & 78.4 & 84.4 
& 14.0 & - \\
SA-COMB~\cite{duarte2022sign} & 55.8 & 79.6 & 92.1 & \textbf{1.0} & - & 53.1 & 79.4 & 86.1 
&\textbf{1.0} & - \\
CiCo~\cite{cheng2023cico} & 69.5& 86.6 & 92.1 & \textbf{1.0} & - & 70.2 & 88.0 & 92.8 
&\textbf{1.0} & - \\
CiCo*~\cite{cheng2023cico} & 70.4 & 88.2 & 92.7 & \textbf{1.0} & 4.8 & 70.9 & 87.2 & 92.5 
&\textbf{1.0} & 5.1 \\
\midrule
\rowcolor{aliceblue}
\textbf{UPRet} & \textbf{72.0} & \textbf{89.1} & \textbf{94.1} & \textbf{1.0} & \textbf{4.4} & \textbf{72.0} & \textbf{89.4} & \textbf{93.3}
& \textbf{1.0} & \textbf{4.6} \\
\bottomrule
\end{tabular}
\vspace{5pt}
\vspace{-20pt}
\label{tab:comparison sota ph}
\end{table}

\begin{table}[t]
\scriptsize
\centering
\caption{\textbf{Comparison with the SOTA on CSL-Daily dataset.} ``$\uparrow$'' denotes that higher is better. ``$\downarrow$'' denotes that lower is better. * denotes our re-implementation of baselines.}
\vspace{-10pt}
\begin{tabular}{lcccccccccc}
\toprule
\multirow{2}{*}{Model} & \multicolumn{5}{c}{T2V} & \multicolumn{5}{c}{V2T} \\ \cmidrule(lr){2-6} \cmidrule(lr){7-11}
  & R@1$\uparrow$ & R@5$\uparrow$ & R@10$\uparrow$ & MedR$\downarrow$ & MnR$\downarrow$ & R@1$\uparrow$ & R@5$\uparrow$ & R@10$\uparrow$ & MedR$\downarrow$ & MnR$\downarrow$ \\  \midrule
CiCo~\cite{cheng2023cico} & 75.3 & 88.2 & 91.9 & \textbf{1.0} & - & 74.7 & 89.4 & 92.2 & \textbf{1.0} & - \\
CiCo*~\cite{cheng2023cico} & 76.3 & 88.6 & \textbf{92.1} & \textbf{1.0} & \textbf{5.8} & 73.9 & 87.9 & 92.0 & \textbf{1.0} & 5.7 \\
\midrule
\rowcolor{aliceblue}
\textbf{UPRet} & \textbf{78.4} & \textbf{89.1} & 92.0 &\textbf{1.0} & 6.7 &\textbf{77.0} & \textbf{89.2} & \textbf{92.7} & \textbf{1.0}  & \textbf{5.5} \\
\bottomrule
\end{tabular}
\label{tab:comparison sota csl}
\end{table}

\section{Experiments}
\subsection{Experimental Settings}
\textbf{Dataset.} While our research focuses on the highly challenging How2Sign~\cite{duarte2021how2sign} dataset. We have also conducted experiments on two other well-known datasets, including PHOENIX-2014T~\cite{camgoz2018neural} and CSL-daily~\cite{zhou2021improving}, to comprehensively evaluate the effectiveness of our method. 
\textbf{How2Sign}~\cite{duarte2021how2sign}, a comprehensive multimodal library of American Sign Language (ASL), encompasses over 80 hours of video across 10 categories reflecting daily life scenarios. It includes 31,164 training, 1,740 validation, and 2,356 test videos, each paired with textual cross-references. \textbf{PHOENIX-2014T}~\cite{camgoz2018neural} features sign language videos from German Public Television's weather forecasts, offering video-to-text translations and annotated timelines. It comprises 7,096 training, 519 validation, and 642 test video-text pairs. \textbf{CSL-Daily}~\cite{zhou2021improving} centered on Chinese Sign Language (CSL) for daily communication, features diverse expressions and phrases across various everyday topics. It includes 18,401 training, 1,077 validation, and 1,176 test sample pairs. These datasets each have unique features and difficulties that make them ideal for measuring sign language video recognition and translation technology state-of-the-art.

\textbf{Evalution metrics.}
We adopt recall at rank (R@K), median rank (MdR), and mean rank (MnR) to evaluate the retrieval performance of all datasets.

\textbf{Implementation Details.} For a fair comparison, following previous work~\cite{cheng2023cico} we first pre-train the I3D network on BSL-1K to get the domain-agnostic sign encoder, the domain-aware sign encoder is fine-tuned by the domain-agnostic sign encoder, the whole sign encoder is obtained by weighted addition of the two. We further initialize our visual and language backbone with CLIP (ViT-B/32). The maximum length of video features and text features are set to 64 and 32 respectively. The model was fine-tuned using the Adam optimizer with the learning rate set to 1e-5. We conducted experiments on 4 NVIDIA A100 GPUs with batch size 512 in 200 epochs.

\subsection{Comparision with State-Of-The-Art Methods}
We compare the proposed UPRet with SPOT-ALGN~\cite{duarte2022sign} and CiCo~\cite{cheng2023cico} on three benchmarks. In Tab~\ref{tab:comparison sota h2s}, we demonstrate the retrieval performance on the How2Sign\cite{duarte2021how2sign}. Previous methods\cite{duarte2022sign,cheng2023cico}, constrained by one-to-one matching, fail to address the uncertainties of sign language. In contrast, by using distributional modeling, we represent actions and words as a spectrum of possible values within a distribution, rather than as single, fixed points. By matching between these distributions and employing optimal transport to minimize the distance between distributions, we achieve fine-grained matching. Our model's superior performance in text-to-video and video-to-text retrieval tasks, with a remarkable 2.7\% and 3.1\% R1 improvement respectively, underlines its effectiveness in achieving accurate matches. The reduced MeanR indicates a consistent accuracy in identifying relevant results. The flexibility of distributional modeling allows our method to handle contextual variations in sign language, demonstrating remarkable adaptability and effectiveness on diverse datasets such as PHOENIX-2014T\cite{camgoz2018neural} and CSL-Daily\cite{zhou2021improving}, as shown in Tab.~\ref{tab:comparison sota ph} and Tab.~\ref{tab:comparison sota csl}.


\begin{table}[t]
\centering
\begin{minipage}[t]{0.48\textwidth}
\scriptsize
\centering
\caption{\textbf{Ablation study of each module on How2Sign.}}
\vspace{-10pt}
\begin{tabular}{ccccccc}
\toprule
\multirow{2}{*}{DM} & \multirow{2}{*}{RS} & \multirow{2}{*}{OT} & \multicolumn{4}{c}{\textbf{How2Sign}} \\ \cmidrule(rl){4-7} 
 &  &   & R@1$\uparrow$ & R@5$\uparrow$ & R@10$\uparrow$ & MnR$\downarrow$ \\  \midrule
 & &    & 56.4 & 69.4 & 74.1 & 58.8 \\
 \boldcheckmark & \boldcheckmark & & 56.8 & 69.8 & 74.7 & 60.6  \\
 &  & \boldcheckmark &  57.8 & 70.4 & 75.2 & 56.8 \\
  &  \boldcheckmark & \boldcheckmark & 58.4 & 71.1 & 74.9 & 62.3  \\
 \midrule
\rowcolor{aliceblue!60}
\boldcheckmark & \boldcheckmark & \boldcheckmark & \textbf{59.1} & \textbf{71.5} & \textbf{75.7} & \textbf{54.4}\\
 \bottomrule
\end{tabular}
\vspace{-15pt}

\label{tab:main abl}
\end{minipage}
\hfill
\begin{minipage}[t]{0.48\textwidth}
\scriptsize
\centering
\caption{\textbf{Time consumption on How2Sign dataset.}}
\vspace{-10pt}
\begin{tabular}{lcccccccc}
\toprule
\multirow{2}{*}{Method} & \multicolumn{3}{c}{T2V} & \multicolumn{1}{c}{Iteration} & \multicolumn{1}{c}{Inference}
\\ \cmidrule(lr){2-4} 
 & R1$\uparrow$ & R5$\uparrow$ & R10$\uparrow$ &Time$\downarrow$ & Time$\downarrow$  \\  \midrule
Baseline & 56.4 & 69.4 & 74.1 & \textbf{0.63} & \textbf{2.54} \\
\midrule
\textbf{UPRet} & \textbf{59.1} & \textbf{71.5} & \textbf{75.7} & 0.70 & 2.55 \\
\bottomrule
\end{tabular}

\vspace{-15pt}
\label{tab:time}
\end{minipage}
\end{table}

\subsection{Ablation Study}
We conduct all ablation studies on the most challenging How2Sign\cite{duarte2021how2sign} dataset.

\textbf{Effectiveness of Each Proposed Component.} In this ablation experiment, we evaluate the impact of each proposed component on the performance of sign video retrieval. As shown in Tab.~\ref{tab:main abl}, our model achieves one-to-many modality mapping through distribution modeling (DM) and random sampling (RS), with R1 rising to 56.8\% for T2V and 51.3\% for V2T. When the optimal transport (OT) is introduced, the accuracy of the model in matching the complex diversity of sign language video and text descriptions is significantly improved, highlighting the importance of OT in feature space alignment. After combining DM, OT, and RS, our method shows significant performance improvements in all evaluation metrics. 


\textbf{Effectiveness of Distribution Modeling.} In Tab.~\ref{tab:ablation dm}, to demonstrate the effectiveness of our distributional modeling, we first modeled the video and text modalities separately, and the R1 metrics for the T2V task decreased by 0.8\% and 1.2\%, respectively, reflecting the inadequacy of single-modality modeling in capturing the fine-grained information required for sign language video retrieval. The performance drop is more pronounced with text-only modeling, highlighting the importance of visual information for detailed representation in sign language video.
In Tab.~\ref{tab:ablation dm structure}, we analyze the specific impact of the different components of the distribution modeling architecture. When using the MLP alone, we noticed a slight drop in performance, suggesting its inadequacy for complex sign language. The introduction of the attention mechanism results in a significant performance boost, emphasizing its effectiveness in capturing contextual details in sign language video and improving model sensitivity. The fusion of MLP and attention mechanisms in our method significantly enhances T2V and V2T tasks, demonstrating their combined strength in navigating the complexity, diversity, and uncertainty of sign language video data.



\begin{table*}[t]
\scriptsize
\begin{minipage}[t]{0.46\textwidth}
\centering
\setlength{\tabcolsep}{4pt}
\caption{\textbf{Ablation study of Distribution Modeling}}
\vspace{-10pt}
\begin{tabular}{cccccc}
\toprule
\multirow{2}{*}{Text} & \multirow{2}{*}{Video} & \multicolumn{4}{c}{T2V} \\ 
\cmidrule(lr){3-6} 
 & & R1$\uparrow$ & R5$\uparrow$ & R10$\uparrow$ & MnR$\downarrow$  \\
\midrule
& & 56.4 & 69.4 & 74.1 & 58.8  \\
\boldcheckmark & & 57.9 & 70.4 & 75.9 & 62.4  \\
& \boldcheckmark & 58.3 & 71.1 & 75.8 & 59.0  \\
\midrule
\rowcolor{aliceblue}
\boldcheckmark & \boldcheckmark & \textbf{59.1} & \textbf{71.5} & \textbf{75.7} & \textbf{54.4} \\
\bottomrule
\end{tabular}
\label{tab:ablation dm}
\end{minipage}
\hfill
\begin{minipage}[t]{0.51\textwidth}
\scriptsize
\caption{\textbf{Ablation study of Distribution Modeling Structures}}
\vspace{-10pt}
\centering
\setlength{\tabcolsep}{4pt}
\begin{tabular}{lcccc}
\toprule
\multirow{2}{*}{Method} & \multicolumn{4}{c}{T2V}  \\ 
\cmidrule(lr){2-5} 
 & R1$\uparrow$ & R5$\uparrow$ & R10$\uparrow$ & MnR$\downarrow$  \\
\midrule
Baseline & 56.4 & 69.4 & 74.1 & 58.8  \\
MLP & 55.2 & 68.7 & 73.1 & 59.3  \\
Attention & 57.9 & 69.9 & 74.3 & 62.4 \\
\midrule
\rowcolor{aliceblue}
\textbf{MLP+Attention} & \textbf{59.1} & \textbf{71.5} & \textbf{75.7} & \textbf{54.4}  \\
\bottomrule
\end{tabular}
\label{tab:ablation dm structure}
\end{minipage}
\end{table*}

\begin{figure}[t] 
\vspace{-5pt}
    \begin{subfigure}{6cm}
        \centering          
        \includegraphics[scale=0.25]{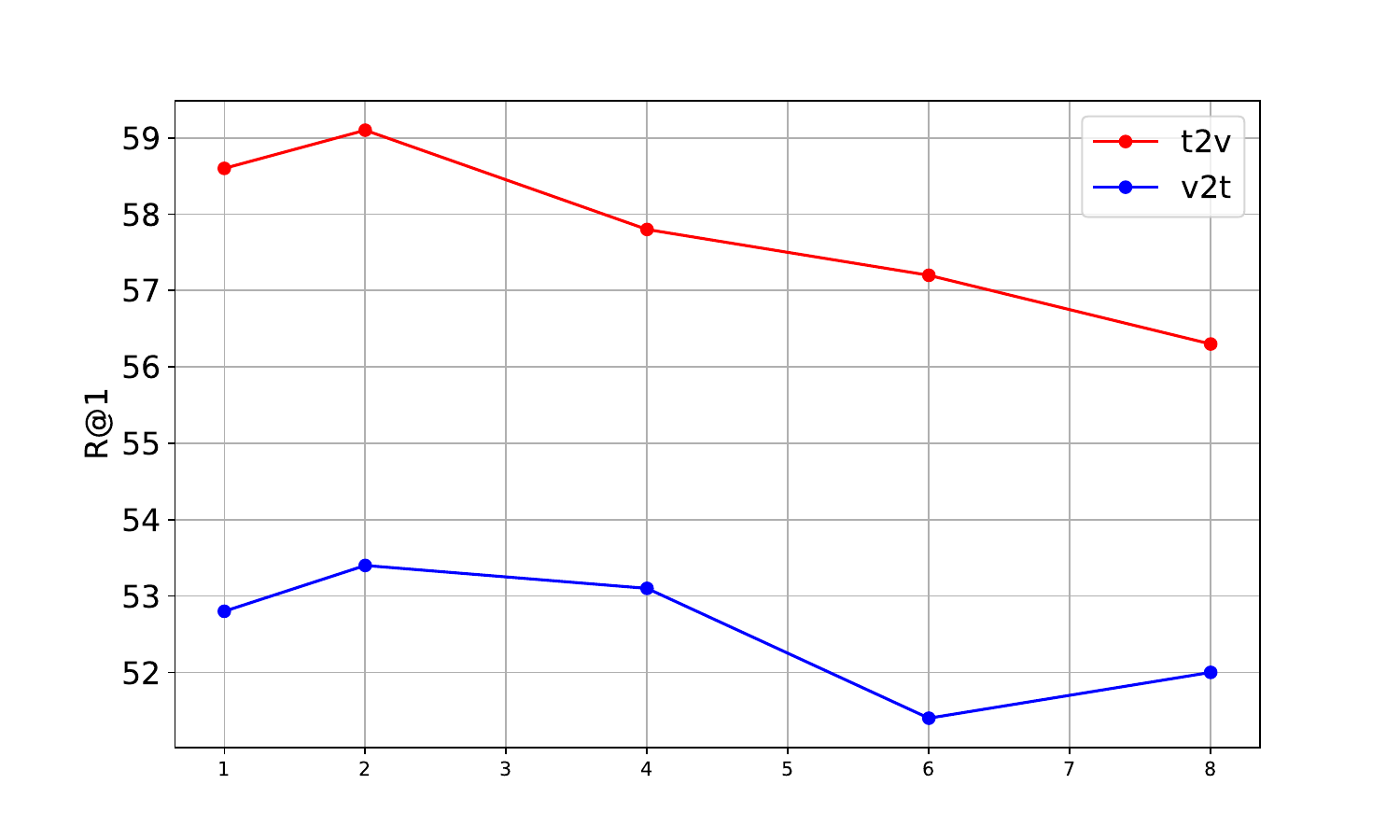}   
        \caption{hyperparameter Sample number} 
        \label{fig:subfig1}
    \end{subfigure}
    \hfill
    \begin{subfigure}{6cm}
        \centering      
        \includegraphics[scale=0.25]{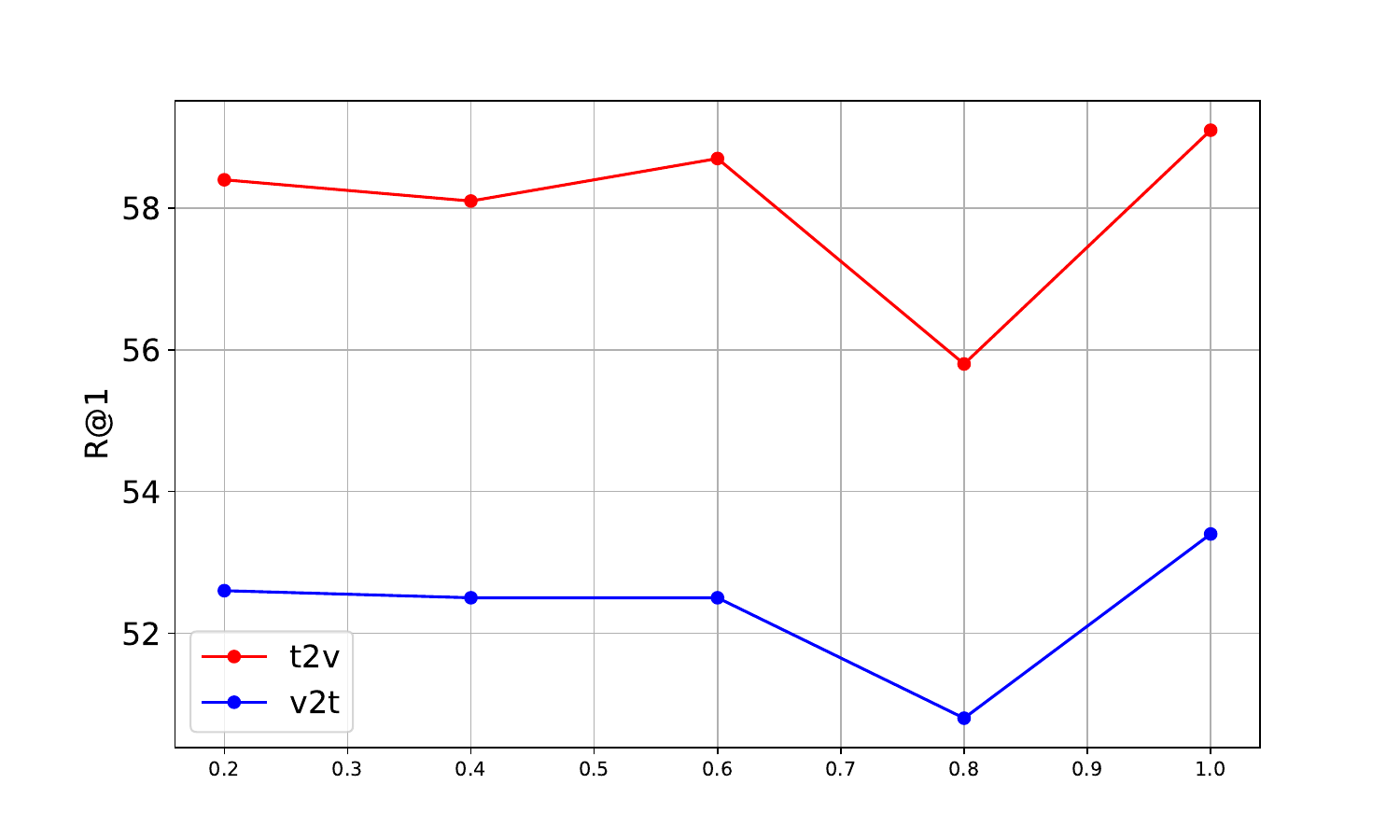}   
        \caption{hyperparameter $\lambda_{ot}$} 
        
        \label{fig:subfig2}
    \end{subfigure}
    \captionsetup{justification=centering, singlelinecheck=false}
    \caption{\textbf{Effect of the hyper-parameters on How2Sign dataset}\\the number of samples in distribution modeling and the weights in optimal transport} 
    \vspace{-20pt}
    \label{fig:parm}  
\end{figure}

\textbf{Effectiveness of  Sample Number.} In the experiments exploring the effect of the number of samples on the performance of sign language video retrieval. As shown in Figure~\ref{fig:parm}(a), when the number of samples was reduced the number of samples to 1, the model performance degraded because a single sample could not capture the complexity of the entire distribution, limiting the depth to which the model could understand the sign language video. On the other hand, as the number of samples is increased to 4, 6, or 8, we notice a gradual decrease in performance, a trend that is because the larger number of samples introduces additional variability, which creates difficulties in training the model and leads to an ineffective convergence to the optimal solution during the training. In summary, the number of samples 2 strikes a trade-off between model complexity and performance.

\begin{figure}[t]
\centering
\includegraphics[width=1.0\linewidth]{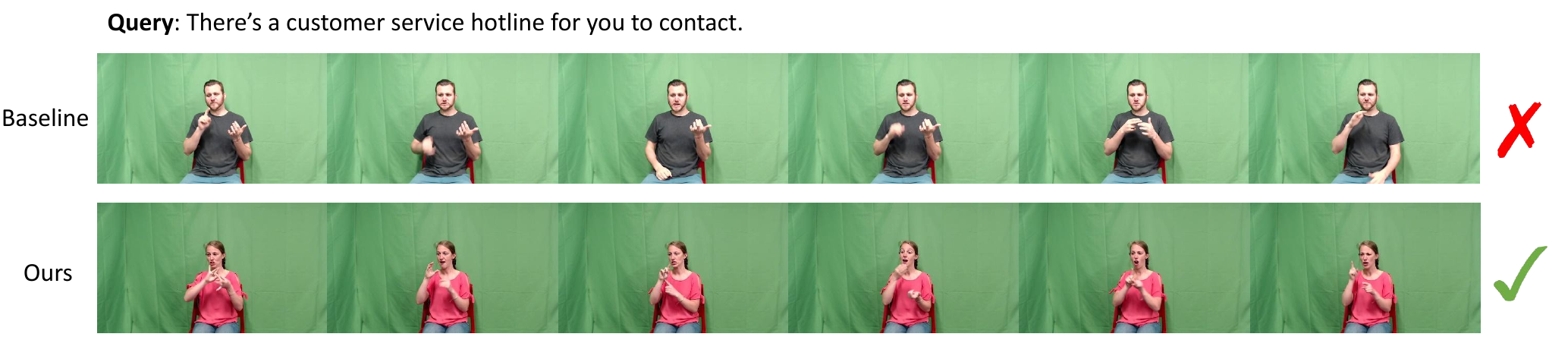}
\vspace{-1em}
\caption{\textbf{Visualization of the text-sign video output on the How2Sign.} {\color{red}Red}: incorrect results of the baseline model. {\color{green}Green}: correct results of our method.}
\vspace{-1em}
\label{fig:output}
\end{figure}


\textbf{Parameter Sensitivity.} The parameter $\lambda_{ot}$ is hyperparameter that trades off $\mathcal{L}_{\mathcal{S}}$.We conducted an assessment of the scale parameter $\lambda_{ot}$ within the interval [0.2, 1], as depicted in Figure~\ref{fig:parm} (b). Based on the insights from Figure~\ref{fig:parm} (b), we determined that setting $\lambda_{ot}$ to 1.0 yields optimal results.

\textbf{Efficiency of Our Method.} In Tab.~\ref{tab:time}, We measure the average training time per iteration and the total inference time. The integration of distributional modeling and OT increases training time, but our method maintains baseline inference times as these modules are inactive during inference.

\section{Conclusion}
In this paper, we creatively define the cross-modal retrieval task as a distribution matching by modeling video and text as probability distributions. Specifically, we achieve this by modeling the representations as distributions to simulate the uncertainty inherent in sign language video and calculating the minimum cost of ensuring modal alignment through OT. Although fine-grained action-word annotations are inaccessible, our method provides new ideas for fine-grained mapping between sign language and natural language and formulates a 1-to-many mapping. Extensive experiments and remarkable performance demonstrate the effectiveness of our method. However, our methods have limitations in cross-lingual sign language retrieval. In future work, we plan to explore more complex uncertainty relations to represent cross-linguistic and cross-cultural differences.

%
%
\bibliographystyle{splncs04}
\bibliography{egbib}
\end{document}